\documentclass[11pt,a4paper,table,dvipsnames]{article}
\usepackage{times,latexsym}
\usepackage{url}
\usepackage[T1]{fontenc}

\newif\ifreview
\reviewfalse

\ifreview
\usepackage[]{tacl2021v1}
\else
\usepackage[acceptedWithA]{tacl2021v1}
\fi

\usepackage{xspace,mfirstuc,tabulary}

\newif\iftaclinstructions
\taclinstructionsfalse %
\iftaclinstructions
\renewcommand{\confidential}{}
\renewcommand{\anonsubtext}{(No author info supplied here, for consistency with
TACL-submission anonymization requirements)}
\newcommand{\instr}
\fi

\iftaclpubformat %

\else

\fi

\usepackage{times}
\usepackage{latexsym}

\usepackage[T1]{fontenc}

\usepackage[utf8]{inputenc}

\usepackage{microtype}

\usepackage{inconsolata}

\usepackage{graphicx}

\usepackage{subcaption}
\usepackage{wrapfig}
\usepackage{booktabs}
\usepackage{amsfonts}
\usepackage{multirow}
\usepackage{longtable}
\usepackage{pifont}
\usepackage{lipsum}
\usepackage{mdframed}
\usepackage{soul}  %

\usepackage{tikz}
\usepackage[edges]{forest}
\definecolor{hidden-draw}{RGB}{0,0,0}

\usepackage{enumitem}
\setlist[itemize]{leftmargin=*,itemsep=0pt,topsep=3pt,parsep=3pt,partopsep=3pt} %

\usepackage{array}
\newcolumntype{M}[1]{>{\centering\arraybackslash}m{#1}}
\newcolumntype{L}[1]{>{\raggedright\arraybackslash}m{#1}}
\newcolumntype{R}[1]{>{\raggedleft\arraybackslash}m{#1}}

\newcommand{\cmark}{\ding{51}}
\newcommand{\xmark}{\ding{55}}

\usepackage{etoolbox}
\AtBeginEnvironment{quote}{\small}

\usepackage[textsize=scriptsize]{todonotes}  %
\newcommand{\rqone}{\mbox{\hyperref[claim-a]{RQ1}}}
\newcommand{\rqtwo}{\mbox{\hyperref[claim-b]{RQ2}}}
\newcommand{\rqthree}{\mbox{\hyperref[claim-c]{RQ3}}}

\newcommand{\psu}{$^1$}
\newcommand{\uiuc}{$^2$}

\title{When Can LLMs {\it Actually} Correct Their Own Mistakes?\\ A Critical Survey of Self-Correction of LLMs}

\author{Ryo Kamoi\psu\quad Yusen Zhang\psu\quad Nan Zhang\psu\quad Jiawei Han\uiuc\quad Rui Zhang\psu \\
  \psu{}Penn State University \quad \uiuc{}University of Illinois Urbana-Champaign \\
  \texttt{\{ryokamoi,rmz5227\}@psu.edu} \\}

\begin{document}
\maketitle

\begin{abstract}
Self-correction is an approach to improving responses from large language models (LLMs) by refining the responses using LLMs during inference. Prior work has proposed various self-correction frameworks using different sources of feedback, including self-evaluation and external feedback. However, there is still no consensus on the question of {\it when LLMs can correct their own mistakes}, as recent studies also report negative results. In this work, we critically survey broad papers and discuss the conditions required for successful self-correction. We first find that prior studies often do not define their research questions in detail and involve impractical frameworks or unfair evaluations that over-evaluate self-correction. To tackle these issues, we categorize research questions in self-correction research and provide a checklist for designing appropriate experiments. Our critical survey based on the newly categorized research questions shows that (1)~no prior work demonstrates successful self-correction with feedback from prompted LLMs, except for studies in tasks that are exceptionally suited for self-correction, (2)~self-correction works well in tasks that can use reliable external feedback, and (3)~large-scale fine-tuning enables self-correction.
\end{abstract}

\section{Introduction}

\begin{figure}[t]
    \centering
    \includegraphics[width=\linewidth]{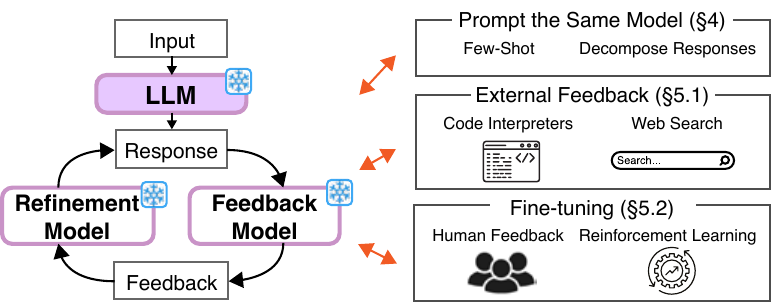}
    \caption{Self-correction in three stages: initial response generation, feedback, and refinement.}
    \label{fig:first_figure}
\end{figure}

Self-correction is a popular approach to improve responses from large language models (LLMs) by refining them using LLMs during inference \citep{bai2022constitutional, madaan2023selfrefine}. Extensive studies on self-correction have been conducted in various tasks, including arithmetic reasoning, code generation, and question answering \citep{gao-etal-2023-rarr, shinn2023reflexion}.
The simplest approach of self-correction prompts LLMs to provide feedback on their own responses and refine the responses using the feedback \citep{huang2024large}, under the hypothesis that {\it recognizing errors is easier than avoiding them} \citep{saunders2022selfcritiquing}. As in Figure~\ref{fig:first_figure}, self-correction has also been studied using additional information for improving feedback, including external tools such as code interpreters \citep{chen2024teaching, gou2024critic}, external knowledge retrieved via web search \citep{gao-etal-2023-rarr, jiang-etal-2023-active}, or fine-tuning \citep{welleck2023generating, selfee2023}. However, recent studies also report negative results indicating that LLMs cannot self-correct \citep{huang2024large, gou2024critic, li-etal-2024-hindsight, chen-etal-2024-tree} or even self-detect \citep{chen2024can, tyen2024llms, hong2024closer, jiang2024selfincorrect, kamoi2024evaluating} their own mistakes at least in certain conditions.
These conflicting observations indicate that further analysis of self-correction is needed.

In this work, we provide a critical survey to investigate the conditions required for successful self-correction. First, our analysis finds that prior studies often do not define their research questions in detail. As a result, many papers fail to provide appropriate experiments to evaluate the research questions they implicitly target. To address this issue, we categorize research questions in self-correction research (\S\ref{sec:claims}) and discuss frameworks that should be used for verifying each research question (\S\ref{sec:framework-categories}). Finally, we provide a checklist for designing appropriate experiments (\S\ref{sec:checklist}).

Next, we analyze prior work to identify when LLMs can self-correct their mistakes, using the new definitions of the research questions. Our analysis highlights that the bottleneck is in the feedback generation~(\S\ref{sec:conclusions-of-analysis}). Specifically, (1)~no prior work shows successful self-correction with feedback from prompted LLMs in general tasks~(\S\ref{sec:prompting-self-correction}), (2)~self-correction works well in tasks where reliable external feedback is available~(\S\ref{sec:external-feedback}), (3)~large-scale fine-tuning enables self-correction~(\S\ref{sec:fine-tuning}), and (4) some tasks have properties exceptionally suitable for self-correction~(\S\ref{sec:prompting-self-correction}).
In summary, our analysis identifies the properties required for successful self-correction as follows:

\vspace{6pt}
\noindent
[\rqone]~When can LLMs self-correct {\it based solely on the inherent capabilities of LLMs?}
\begin{itemize}
    \item In general tasks, no prior work shows reliable evidence of successful self-correction with in-context learning. (\S\ref{sec:prompting-self-correction})
    \item In tasks with specific properties that are exceptionally favorable for self-correction (e.g., responses are decomposable), self-correction is effective even with in-context learning. (\S\ref{sec:prompting-self-correction})
\end{itemize}

\vspace{2pt}
\noindent
[\rqtwo]~When can LLMs self-correct the best-possible initial responses {\it with external information?}
\begin{itemize}
    \item Self-correction is effective in tasks where reliable external feedback is available. (\S\ref{sec:external-feedback})
    \item Fine-tuning enables self-correction when large training data is available but is unexplored for small training data. (\S\ref{sec:fine-tuning})
\end{itemize}

\vspace{2pt}
\noindent
[\rqthree]~When are the final outputs of self-correction {\it better than other approaches}?
\begin{itemize}
    \item Self-correction is often not compared with sufficiently strong baselines, and it is still unclear whether it is better than other approaches. (\S\ref{sec:comparable-baselines})
\end{itemize}\vspace{5pt}

This survey is organized as follows. Section~\ref{sec:overview} provides an overview of self-correction. Section~\ref{sec:target-claims} introduces a new approach to classify research questions and frameworks in self-correction research. Sections~\ref{sec:prompting-self-correction} and \ref{sec:external-info} analyze prior work in self-correction with in-context learning and external information (external tools, external knowledge, fine-tuning), respectively. Section~\ref{sec:comparable-baselines} explains related approaches that should be compared with self-correction as baselines. Section~\ref{sec:conclusions-of-analysis} summarizes our findings from the analysis. Section~\ref{sec:checklist} provides a checklist for self-correction research. Section~\ref{sec:other-survey} explains differences from other surveys. Section~\ref{sec:related-work} provides studies related to self-correction. Section~\ref{sec:future-directions} provides future directions.

\paragraph{Timeframe.}
This survey was originally published in June 2024 and primarily includes research papers and studies published up to and including May 2024. While papers published after this date are not comprehensively analyzed, they are briefly discussed in Section~\ref{sec:recent-papers}.

\let\oldcolorbox\colorbox
\renewcommand{\colorbox}{\setlength{\fboxsep}{0pt}\oldcolorbox}
\newcommand{\posthoc}{Post-hoc}
\newcommand{\gentime}{\textcolor{red}{Gen Time}}
\newcommand{\oracle}{\textcolor{black}{Oracle}}
\newcommand{\intrinsic}{\textcolor{blue}{Intrinsic}}
\newcommand{\fairasymmetric}{\textcolor{Green}{Fair-Asym.}}
\newcommand{\unfairasymmetric}{\textcolor{darkgray}{Unfair-Asym.}}
\newcommand{\crossmodel}{\textcolor{orange}{Cross-Model}}
\begin{table*}[t]
    \setlength{\tabcolsep}{2pt}
    \renewcommand{\arraystretch}{.5}
    \newcommand{\ccmidrule}{\cmidrule{1-11}}
    \newcommand{\tworows}[1]{\multirow{3.7}{\linewidth}{\centering #1}}
    \newcommand{\claimrow}[1]{\multicolumn{11}{c}{\scriptsize #1}}
    \tiny
    \centering
    \begin{tabular}{M{.1\linewidth}M{.06\linewidth}M{.1\linewidth}
    M{.04\linewidth}M{.08\linewidth}M{.095\linewidth}
    M{.07\linewidth}M{.09\linewidth}M{.1\linewidth}M{.09\linewidth}M{.065\linewidth}}
        \toprule
        \tworows{Paper} & \tworows{Category} & \tworows{Main Models} & \multicolumn{3}{c}{Additional Feedback} & \multicolumn{5}{c}{Main Tasks} \\
        \cmidrule(l{2pt}r{2pt}){4-6} \cmidrule(l{2pt}r{2pt}){7-11}
        &&& Oracle & External Tools & Fine-Tuning & Reasoning, Coding & Closed-book, Knowledge & Open-book, Context-based & Open-ended Text~Gen & Decom-posable \\
        \midrule
        \claimrow{Self-Correction with {\bf In-context Learning} ({\it Intrinsic Self-Correction})} \\
        \midrule
        \phantom{-----}CoVe\phantom{-----} \citeyearpar{dhuliawala2023chainofverification} &
        \intrinsic &  PaLM 540B &
        -- & -- & -- & -- & -- & -- & -- & Multiple Answers \\
        \ccmidrule
        \cellcolor[RGB]{211,211,211}{CAI Revisions \citeyearpar{bai2022constitutional}$^{\spadesuit}$} &
        \intrinsic & 52B (no~details) &
        -- & -- & -- & -- & -- & -- & Detoxification & -- \\
        \ccmidrule
        \cellcolor[RGB]{211,211,211}{Self-Refine \citeyearpar{madaan2023selfrefine}$^{\spadesuit}$} &
        \intrinsic & GPT-3.5, GPT-4 &
        -- & -- & -- & Math, Coding & -- & Dialogue & -- & -- \\
        \ccmidrule
        \phantom{-------}RCI\phantom{-------} \citeyearpar[\S3.1]{kim2023computer} &
        \oracle & GPT-3.5-T &
        \cellcolor[RGB]{211,211,211}{\checkmark} & -- & -- & Computer Tasks & CSQA & -- & -- & -- \\
        \ccmidrule
        \phantom{----}Reflexion\phantom{----} \citeyearpar[\S4.2]{shinn2023reflexion} &
        \oracle & GPT-4 &
        \cellcolor[RGB]{211,211,211}{\checkmark} & -- & -- & -- & -- & HotpotQA (GT~Context) & -- & -- \\
        \midrule
        \claimrow{Self-Correction with {\bf External Tools or Knowledge}} \\
        \midrule
        \phantom{----}Reflexion\phantom{----} \citeyearpar[\S4.1, 4.3]{shinn2023reflexion} &
        \fairasymmetric & GPT-4 &
        -- & Game Envs, Interpreter & -- & Games, Coding & -- & -- & -- & -- \\
        \ccmidrule
        Self-Debug \citeyearpar{chen2024teaching} &
        \fairasymmetric & GPT-3.5-T, GPT-4 &
        -- & Code Interpreter & -- & Text-to-Code & -- & -- & -- & -- \\
        \ccmidrule
        \phantom{----}CRITIC\phantom{----} \citeyearpar{gou2024critic} &
        \fairasymmetric & GPT-3, Llama~2~70B &
        -- & Interpreter, Web Search & -- & GSM8k, SVAMP & HotpotQA & -- & Detoxification & -- \\
        \ccmidrule
        \phantom{----}RARR\phantom{----} \citeyearpar{gao-etal-2023-rarr} &
        \unfairasymmetric & Palm 540B &
        -- & Web Search & -- &
        -- & NQ, SQA, QReCC & -- & -- & -- \\
        \ccmidrule
        \phantom{----}Reflexion\phantom{----} \citeyearpar[\S4.2]{shinn2023reflexion} &
        \oracle & GPT-4 &
        \cellcolor[RGB]{211,211,211}{\checkmark} & Wikipedia API & -- &
        -- & HotpotQA & -- & -- & -- \\
        \midrule
        \claimrow{Self-Correction with {\bf Fine-tuning}} \\
        \midrule
        Self-Critique \citeyearpar{saunders2022selfcritiquing} &
        \fairasymmetric & InstructGPT &
        -- & -- & Human Assessment & -- & -- & Topic-based Summarization & -- & -- \\
        \ccmidrule
        \phantom{-----}SelFee\phantom{-----} \citeyearpar{selfee2023} &
        \fairasymmetric & Llama 7B,~13B &
        -- & -- & ChatGPT Assessment & MT-Bench & MT-Bench & MT-Bench & MT-Bench & -- \\
        \ccmidrule
        \phantom{-----}Baldur\phantom{-----} \citeyearpar{first2023baldur} &
        \fairasymmetric & Minerva 8B~,62B &
        -- & Proof Assistant & GT Answer & Proof Generation & -- & -- & -- & -- \\
        \ccmidrule
        \phantom{---}REFINER\phantom{---} \citeyearpar{paul-etal-2024-refiner} &
        \crossmodel & GPT-3.5 (FB:T5-base) &
        -- & -- & Synthetic Data & Math, Logic & -- & -- & Moral Stories & -- \\
        \ccmidrule
        RL4F \phantom{-----}\citeyearpar{akyurek-etal-2023-rl4f}\phantom{-----} &
        \crossmodel & GPT-3 (FB:~T5-large) &
        -- & -- & Reinforcement Learning & Action Planning & -- & Topic-based Summarization & -- & -- \\
        \ccmidrule
        Self-Correction \citeyearpar[\S3.4]{welleck2023generating} &
        \crossmodel & GPT-3 (FB:~GPT-Neo) &
        -- & -- & GT Answer, External & GSM8k, SVAMP & -- & -- & Detoxification & -- \\
        \ccmidrule
        Self-Correction \citeyearpar[\S3.1-3.3]{welleck2023generating} &
        \unfairasymmetric & GPT-Neo 1.3B, GPT-2 &
        -- & -- & GT Answer, External & GSM8k, SVAMP & -- & -- & Detoxification, Const Gen & -- \\
        \midrule
        \claimrow{\textcolor{red}{Negative Results} of Self-Correction (i.e., LLMs cannot Self-Correct)} \\
        \midrule
        RCI (Table~17) \citeyearpar{kim2023computer}  &
        \intrinsic & GPT-3.5-T &
        -- & -- & -- & Computer Tasks & CSQA & -- & -- & -- \\
        \ccmidrule
        CRITIC~w/o~Tool \citeyearpar{gou2024critic} &
        \intrinsic & GPT-3, Llama~2~70B &
        -- & -- & -- & GSM8k, SVAMP & Closed-book HotpotQA & -- & Detoxification & -- \\
        \ccmidrule
        \phantom{---}\citeauthor{huang2024large}\phantom{---} \citeyearpar{huang2024large} &
        \intrinsic & GPT-4-T, GPT-3.5-T &
        -- & -- & -- & GSM8k & CSQA, HotpotQA & -- & -- & -- \\
        \bottomrule
    \end{tabular}
    \caption{Representative studies in self-correction of LLMs. \colorbox[RGB]{211,211,211}{Gray color} represents unrealistic settings. $^{\spadesuit}$:~Weak prompts for generating initial responses. FB: Feedback models for cross-model correction.} \label{tab:priorwork}
\end{table*}

\begin{figure*}[t]
    \centering
    \includegraphics[width=\linewidth]{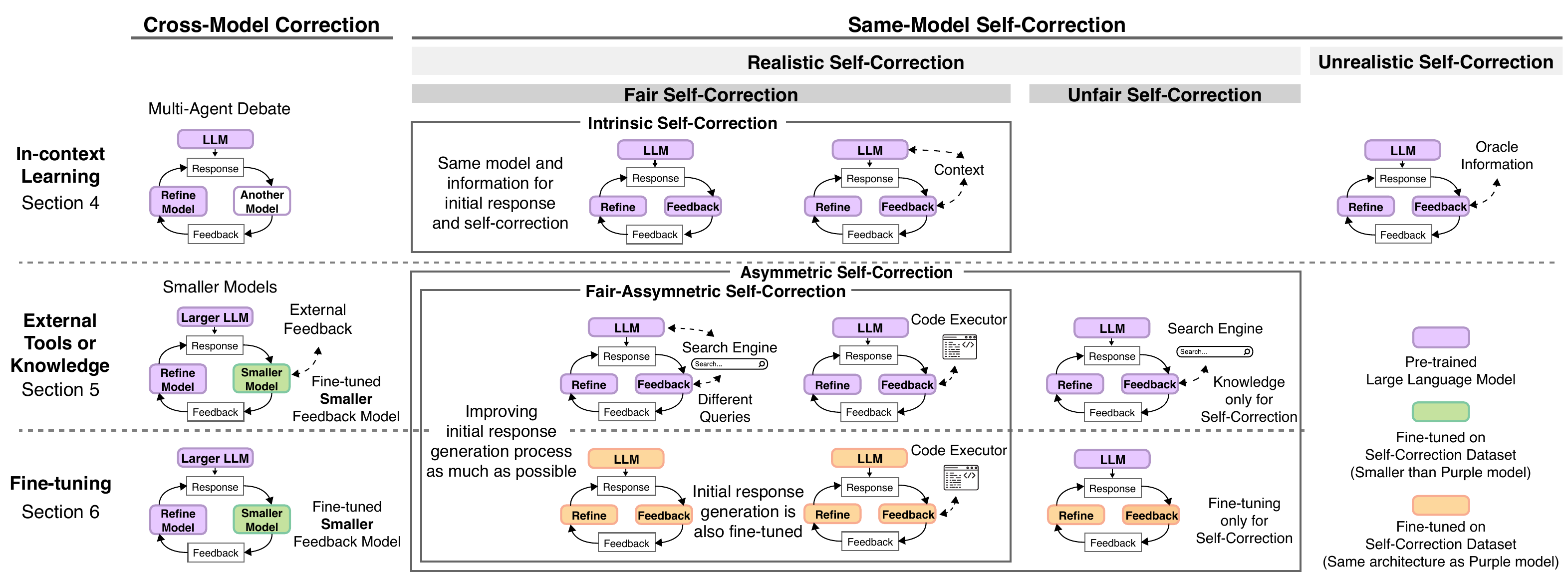}
    \caption{LLM self-correction frameworks, categorized by information used for generating feedback and whether they use best-possible initial responses (\S\ref{sec:framework-categories}). This figure illustrates representative architectures.}
    \label{fig:categories}
\end{figure*}

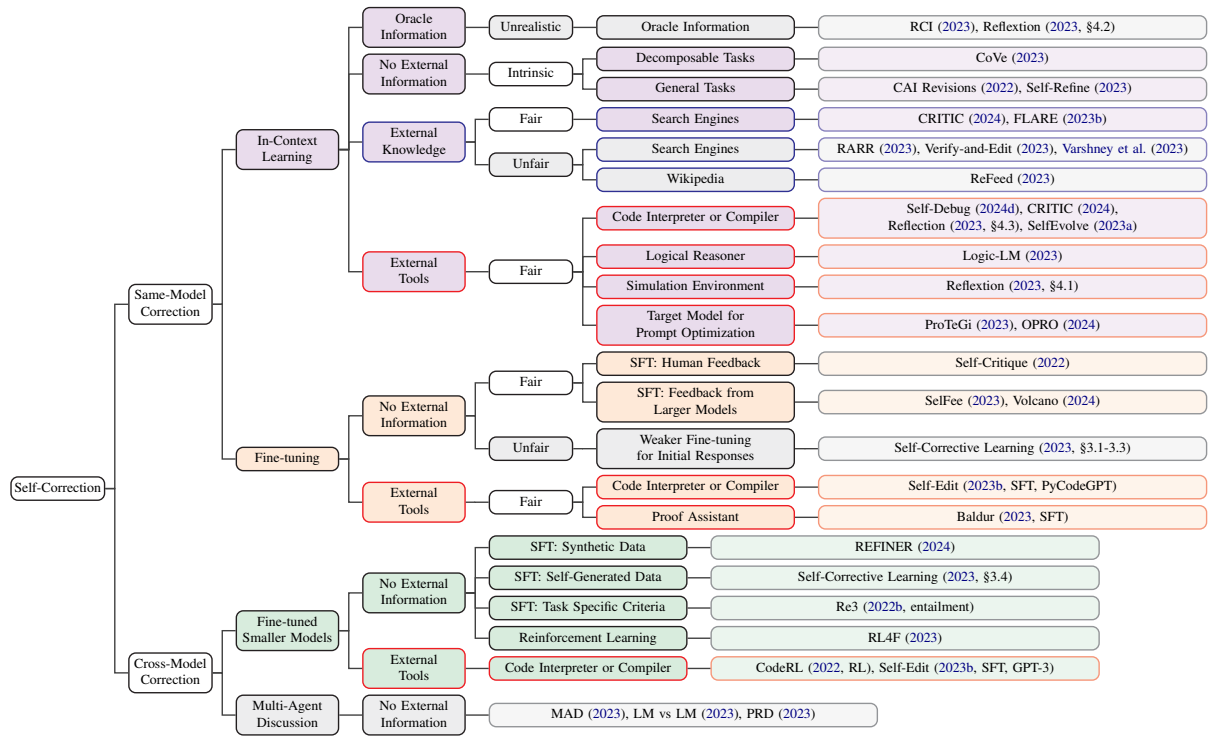
\begin{figure*}[t!]
    \centering
    \tikzstyle{my-box}=[
    rectangle,
    rounded corners,
    text opacity=1,
    minimum height=0.1em,
    minimum width=0.1em,
    inner sep=2pt,
    align=center,
    /tikz/align=center,
    fill opacity=.5,
    ]
    \tikzstyle{leaf}=[my-box]
    \resizebox{\textwidth}{!}{
        \newcommand{\twlevelfive}{10em}
        \newcommand{\twlevelsix}{20em}
        \begin{forest}
            forked edges,
            for tree={
                grow=east,
                reversed=true,
                anchor=base west,
                parent anchor=east,
                child anchor=west,
                base=left,
                font=\scriptsize,
                rectangle,
                draw=hidden-draw,
                rounded corners,
                align=center,
                /tikz/align=center,
                minimum width=0.1em,
                edge+={darkgray, line width=0.8pt},
                s sep=2pt,
                inner xsep=2pt,
                inner ysep=2pt,
                outer ysep=1pt,
                ver/.style={rotate=90, child anchor=north, parent anchor=south, anchor=center},
            },
            where level=1{text width=4em,font=\scriptsize}{},
            where level=2{text width=5em,font=\scriptsize}{},
            where level=3{text width=5em,font=\scriptsize}{},
            where level=4{text width=4em,font=\scriptsize}{},
            where level=5{text width=\twlevelfive,font=\scriptsize}{},
            where level=6{text width=\twlevelsix,font=\scriptsize}{},
            [
                Self-Correction, draw=gray, color=Black!100, fill=White!15, thick, text=black, outer ysep=0pt
                [
                    Same-Model \\ Correction
                    , color=Black!100, fill=White!15, thick, text=black, outer ysep=0pt, l=5pt
                    [
                        In-Context \\ Learning
                        , color=Black!100, fill=Purple!15, thick, text=black, outer ysep=0pt, l=5pt
                        [
                            Oracle \\ Information
                            , color=Black!100, fill=Purple!15, thick, text=black, l=5pt
                            [
                                Unrealistic
                                , color=Black!100, fill=Gray!15, thick, text=black,
                                [
                                    Oracle Information
                                    , color=Black!100, fill=Gray!15, thick, text=black
                                    [   
                                        {RCI \citeyearpar{kim2023computer}, Reflextion \citeyearpar[\S4.2]{shinn2023reflexion}}
                                        , leaf, color=Black!50, fill=Gray!15, thick, text=black
                                    ]
                                ]
                            ]
                        ]
                        [
                            No External \\ Information
                            , color=Black!100, fill=Purple!15, thick, text=black, l=5pt
                            [
                                Intrinsic
                                , color=Black!100, fill=White!15, thick, text=black
                                [
                                    Decomposable Tasks
                                    , color=Black!100, fill=Purple!15, thick, text=black
                                    [   
                                        CoVe \citeyearpar{dhuliawala2023chainofverification}
                                        , leaf, color=Black!50, fill=Purple!15, thick, text=black
                                    ]
                                ]
                                [
                                    General Tasks
                                    , color=Black!100, fill=Purple!15, thick, text=black
                                    [
                                        {CAI Revisions \citeyearpar{bai2022constitutional}, Self-Refine \citeyearpar{madaan2023selfrefine}}
                                        , leaf, color=Black!50, fill=Purple!15, thick, text=black
                                    ]
                                ]
                            ]
                        ]
                        [
                            External \\ Knowledge
                            , color=Blue!100, fill=Purple!15, thick, text=black, l=5pt
                            [
                                Fair
                                , color=Black!100, fill=White!15, thick, text=black
                                [
                                    Search Engines
                                    , color=Blue!100, fill=Purple!15, thick, text=black
                                    [
                                        {CRITIC \citeyearpar{gou2024critic}, FLARE \citeyearpar{jiang-etal-2023-active}}
                                        , leaf, color=Blue!50, fill=Purple!15, thick, text=black
                                    ]
                                ]
                            ]
                            [
                                Unfair
                                , color=Black!100, fill=Gray!15, thick, text=black
                                [
                                    Search Engines
                                    , color=Blue!100, fill=Gray!15, thick, text=black
                                    [
                                        {RARR \citeyearpar{gao-etal-2023-rarr}, Verify-and-Edit \citeyearpar{zhao-etal-2023-verify}, \citet{varshney2023stitch}}
                                        , leaf, color=Blue!50, fill=Gray!15, thick, text=black
                                    ]
                                ]
                                [
                                    Wikipedia
                                    , color=Blue!100, fill=Gray!15, thick, text=black
                                    [
                                        {ReFeed \citeyearpar{yu2023improving}}
                                        , leaf, color=Blue!50, fill=Gray!15, thick, text=black
                                    ]
                                ]
                            ]
                        ]
                        [
                            External \\ Tools
                            , color=Red!100, fill=Purple!15, thick, text=black, l=5pt
                            [
                                Fair
                                , color=Black!100, fill=White!15, thick, text=black
                                [
                                    Code Interpreter or Compiler
                                    , color=Red!100, fill=Purple!15, thick, text=black
                                    [
                                        {Self-Debug \citeyearpar{chen2024teaching}, CRITIC \citeyearpar{gou2024critic},\\
                                        Reflection \citeyearpar[\S4.3]{shinn2023reflexion}, SelfEvolve \citeyearpar{jiang2023selfevolve}}
                                        , leaf, color=Red!50, fill=Purple!15, thick, text=black
                                    ]
                                ]
                                [
                                    Logical Reasoner
                                    , color=Red!100, fill=Purple!15, thick, text=black
                                    [
                                        Logic-LM \citeyearpar{pan-etal-2023-logic}
                                        , leaf, color=Red!50, fill=Purple!15, thick, text=black
                                    ]
                                ]
                                [
                                    Simulation Environment
                                    , color=Red!100, fill=Purple!15, thick, text=black
                                    [
                                        {Reflextion \citeyearpar[\S4.1]{shinn2023reflexion}}
                                        , leaf, color=Red!50, fill=Purple!15, thick, text=black
                                    ]
                                ]
                                [
                                    Target Model for \\ Prompt Optimization 
                                    , color=Red!100, fill=Purple!15, thick, text=black
                                    [
                                        {ProTeGi \citeyearpar{pryzant-etal-2023-automatic}, OPRO \citeyearpar{yang2024large}}
                                        , leaf, color=Red!50, fill=Purple!15, thick, text=black
                                    ]
                                ]
                            ]
                        ]
                    ]
                    [
                        Fine-tuning
                        , color=Black!100, fill=Orange!15, thick, text=black, outer ysep=0pt, l=5pt
                        [
                            No External \\ Information
                            , color=Black!100, fill=Orange!15, thick, text=black, l=5pt
                            [
                                Fair
                                , color=Black!100, fill=White!15, thick, text=black
                                [
                                   SFT: Human Feedback,
                                   color=Black!100, fill=Orange!15, thick, text=black
                                   [
                                        Self-Critique \citeyearpar{saunders2022selfcritiquing}
                                        , leaf, color=Black!50, fill=Orange!15, thick, text=black, 
                                   ]
                                ]
                                [
                                   SFT: Feedback from \\ Larger Models,
                                   color=Black!100, fill=Orange!15, thick, text=black
                                   [
                                        {SelFee \citeyearpar{selfee2023}, Volcano \citeyearpar{lee2023volcano}}
                                        , leaf, color=Black!50, fill=Orange!15, thick, text=black, 
                                   ]
                                ]
                            ]
                            [
                                Unfair
                                , color=Black!100, fill=Gray!15, thick, text=black
                                [
                                    Weaker Fine-tuning \\ for Initial Responses,
                                    color=Black!100, fill=Gray!15, thick, text=black
                                    [
                                        {Self-Corrective Learning \citeyearpar[\S3.1-3.3]{welleck2023generating}}
                                        , leaf, color=Black!50, fill=Gray!15, thick, text=black, 
                                    ]
                                ]
                            ]
                        ]
                        [
                            External \\ Tools
                            , color=Red!100, fill=Orange!15, thick, text=black, l=5pt
                            [
                                Fair
                                , color=Black!100, fill=White!15, thick, text=black
                                [   
                                    Code Interpreter or Compiler
                                    , color=Red!100,, fill=Orange!15, thick, text=black, 
                                    [
                                        {Self-Edit \citeyearpar[SFT, PyCodeGPT]{zhang-etal-2023-self}}
                                        , leaf, color=Red!50, fill=Orange!15, thick, text=black
                                    ]
                                ]
                                [   
                                    Proof Assistant
                                    , color=Red!100, fill=Orange!15, thick, text=black, 
                                    [
                                        {Baldur \citeyearpar[SFT]{first2023baldur}}
                                        , leaf, color=Red!50, fill=Orange!15, thick, text=black
                                    ]
                                ]
                            ]
                        ]
                    ]
                ]
                [
                    Cross-Model \\ Correction
                    , color=Black!100, fill=White!15, thick, text=black, outer ysep=0pt, l=5pt
                    [
                        Fine-tuned \\ Smaller Models
                        , color=Black!100, fill=Green!15, thick, text=black, outer ysep=0pt, l=5pt
                        [
                            No External \\ Information
                            , color=Black!100, fill=Green!15, thick, text=black, l=5pt
                                [
                                   SFT: Synthetic Data,
                                   color=Black!100, fill=Green!15, thick, text=black, text width=\twlevelfive
                                   [
                                        REFINER \citeyearpar{paul-etal-2024-refiner}
                                        , leaf, color=Black!50, fill=Green!15, thick, text=black, text width=\twlevelsix
                                   ]
                                ]
                                [
                                   SFT: Self-Generated Data,
                                   color=Black!100, fill=Green!15, thick, text=black, text width=\twlevelfive
                                   [
                                        {Self-Corrective Learning \citeyearpar[\S3.4]{welleck2023generating}}
                                        , leaf, color=Black!50, fill=Green!15, thick, text=black, text width=\twlevelsix
                                   ]
                                ]
                                [
                                   SFT: Task Specific Criteria,
                                   color=Black!100, fill=Green!15, thick, text=black, text width=\twlevelfive
                                   [
                                        {Re3 \citeyearpar[entailment]{yang-etal-2022-re3}}
                                        , leaf, color=Black!50, fill=Green!15, thick, text=black, text width=\twlevelsix
                                   ]
                                ]
                                [
                                   Reinforcement Learning,
                                   color=Black!100, fill=Green!15, thick, text=black, text width=\twlevelfive
                                   [
                                        RL4F \citeyearpar{akyurek-etal-2023-rl4f}
                                        , leaf, color=Black!50, fill=Green!15, thick, text=black, text width=\twlevelsix
                                   ]
                                ]
                        ]
                        [
                            External \\ Tools
                            , color=Red!100, fill=Green!15, thick, text=black, l=5pt
                                [
                                   Code Interpreter or Compiler,
                                   color=Red!100, fill=Green!15, thick, text=black, text width=\twlevelfive
                                   [
                                        {CodeRL \citeyearpar[RL]{le2022coderl}, Self-Edit \citeyearpar[SFT, GPT-3]{zhang-etal-2023-self}}
                                        , leaf, color=Red!50, fill=Green!15, thick, text=black, text width=\twlevelsix
                                   ]
                                ]
                        ]
                    ]
                    [
                        Multi-Agent \\ Discussion
                        , color=Black!100, fill=Gray!15, thick, text=black, outer ysep=0pt, l=5pt
                        [
                            No External \\ Information
                            , color=Black!100, fill=Gray!15, thick, text=black, l=5pt
                           [
                                {MAD \citeyearpar{liang2023encouraging}, LM vs LM \citeyearpar{cohen-etal-2023-lm}, PRD \citeyearpar{li2023prd}}
                                , leaf, color=Black!50, fill=Gray!15, thick, text=black, text width=\twlevelsix
                           ]
                        ]
                    ]
                ]
            ]
        \end{forest}
    }
    \caption{Taxonomy of LLM self-correction, categorized by information used for generating feedback and whether they use best-possible initial responses (fair or unfair). Refer to Section~\ref{sec:framework-categories} for the definitions.}
    \label{fig:papers-categorized}
\end{figure*}

\newcommand{\redxmark}{\textcolor{red}{\xmark}}
\newcommand{\doublegreencmark}{\textcolor{teal}{\cmark\cmark}}
\newcommand{\greencmark}{\textcolor{teal}{\cmark}}
\begin{table*}[t]
    \setlength{\tabcolsep}{7pt}
    \fontsize{7pt}{7pt}\selectfont
    \centering
    \begin{tabular}{M{.05\textwidth}M{.1\textwidth}M{.1\textwidth}M{.1\textwidth}
    M{.1\textwidth}M{.1\textwidth}M{.1\textwidth}M{.1\textwidth}}
    \toprule
        RQ &
        Self-Refine \citeyearpar{madaan2023selfrefine} & \citet{huang2024large} & 
        \phantom{-----}RCI\phantom{-----} \citeyearpar[\S3.1]{kim2023computer} & \phantom{-----}RCI\phantom{-----} \citeyearpar[\S3.2]{kim2023computer} &
        CRITIC \mbox{\citeyearpar[\S4.2]{gou2024critic}} & CRITIC \mbox{\citeyearpar[\S4.3]{gou2024critic}} &
        \phantom{-----}RARR\phantom{-----} \citeyearpar{gao-etal-2023-rarr} \\
    \midrule
        \rqone &
        \greencmark & \redxmark{} (\S3,5)  & 
        \greencmark & --          &  %
        \redxmark   & \redxmark   &  %
        --          \\
        \rqtwo &
        --          & --          &
        --          & \greencmark &
        \greencmark & \greencmark &
        -- \\
        \rqthree &
        --          & \redxmark{} (\S4) &
        --          & \greencmark &
        --          & \greencmark &
        \greencmark \\
    \bottomrule
    \end{tabular}
    \caption{Research questions that prior studies implicitly target by claiming they are \greencmark{} verified or \redxmark{} refuted.}
    \label{tab:claims-in-prior-work}
\vskip 1em
    \setlength{\tabcolsep}{5pt}
    \fontsize{7pt}{7pt}\selectfont
    \centering
    \begin{tabular}{M{.04\textwidth}M{.17\textwidth}M{.19\textwidth}M{.06\textwidth}
    M{.2\textwidth}M{.2\textwidth}}
    \toprule
        \multirow{2.7}{*}{RQ} & \multicolumn{3}{c}{Requirements for Frameworks} & \multicolumn{2}{c}{Required Experiments} \\
        \cmidrule(l{2pt}r{2pt}){2-4} \cmidrule(l{2pt}r{2pt}){5-6}
        & Information Symmetricity & Best-possible Initial Responses & Realistic
        & Comparison to Initial Responses & Comparison to Strong Baselines \\
    \midrule
        \rqone &
        \checkmark & \checkmark & \checkmark &
        \checkmark & -- \\
        \rqtwo &
        --         & \checkmark & \checkmark &
        \checkmark & -- \\
        \rqthree &
        --         & --         & \checkmark &
        --         & \checkmark \\
    \bottomrule
    \end{tabular}
    \caption{Requirements for experiments to verify each research question in Section~\ref{sec:claims}.}
    \label{tab:claims-requirements}
\end{table*}

\section{Self-Correction of LLMs} \label{sec:overview}
The term ``self-correction'' is used in a wide range of scenarios, from a strict definition in which LLMs refine their own responses by themselves \citep{madaan2023selfrefine, huang2024large} to broader concepts that also involve feedback from external tools or knowledge \citep{shinn2023reflexion, gou2024critic}. In this work, we define self-correction as a framework that {\it refines} responses from LLMs using LLMs {\it during inference}, possibly with external tools or knowledge. As in Table~\ref{tab:priorwork}, Figure~\ref{fig:categories}, and Figure~\ref{fig:papers-categorized}, self-correction has been studied in various frameworks with different sources of feedback.

\subsection{Frameworks}

Prior studies propose self-correction frameworks with various different architectures.

\paragraph{Explicit Feedback vs. Direct Refinement.}
Self-correction often consists of three stages including {\it feedback generation} \citep{kim2023computer, madaan2023selfrefine, shinn2023reflexion, huang2024large}:

\begin{itemize}
    \item {\bf Initial Response Generation} is a stage of generating initial responses from an LLM.
    \item {\bf Feedback} model generates feedback given the original input and initial response. This stage may use external tools or knowledge.
    \item {\bf Refinement} model generates a refined response, given the input, initial response, and feedback.
\end{itemize}

\noindent
{\it Direct refinement} is another approach that refines responses without generating feedback explicitly \citep{saunders2022selfcritiquing, bai2022constitutional, welleck2023generating, akyurek-etal-2023-rl4f}.

\paragraph{Post-hoc vs. Generation-time.}
{\it Post-hoc correction} refines responses after they are generated \citep{pan2024automatically}.
{\it Generation-time correction} or step-level correction \citep{paul-etal-2024-refiner, jiang-etal-2023-active} improves step-by-step reasoning by providing feedback on intermediate reasoning steps. Post-hoc correction is more flexible and applicable to broader tasks, although generation-time correction is popular for reasoning tasks \citep{pan2024automatically}.

\paragraph{Same-model vs. Cross-model.}
{\it Cross-model correction} generates feedback or refines the responses using models different from the model that generates initial responses. Cross-model correction has been mostly studied in the settings of correcting mistakes of large proprietary LLMs using small fine-tuned models \citep{welleck2023generating, akyurek-etal-2023-rl4f, paul-etal-2024-refiner} or multi-agent debate of multiple models with similar capabilities \citep{liang2023encouraging, li2023prd, cohen-etal-2023-lm, du2023improving, Zhang2023ExploringCM, chen2024reconcile, chan2024chateval, Wang2024RethinkingTB}.

\subsection{Sources of Feedback}
\paragraph{Intrinsic ({\normalfont \S\ref{sec:prompting-self-correction}}).}
Intrinsic self-correction prompts LLMs to generate feedback on their own responses. Prompting strategies include simple zero-shot or few-shot prompts \citep{madaan2023selfrefine, kim2023computer}, decomposing the responses \citep{dhuliawala2023chainofverification}, and evaluating confidence \citep{varshney2023stitch, jiang-etal-2023-active, wu2024large}.

\paragraph{External Information ({\normalfont \S\ref{sec:external-feedback}}).}
Self-correction often relies on external information, including {\bf external tools} such as code executors \citep{jiang2023selfevolve, gou2024critic, chen2024teaching, stengeleskin2024regal}, symbolic reasoners \citep{pan-etal-2023-logic}, proof assistant \citep{first2023baldur}, or task-specific metrics \citep{xu2023instructscore}, {\bf external knowledge} from search engines \citep{jiang-etal-2023-active, gao-etal-2023-rarr, zhao-etal-2023-verify}, Wikipedia \citep{yu2023improving, zhao-etal-2023-verify}, or other corpora \citep{peng2023check, zhao-etal-2023-verify}, {\bf oracle information} such as ground-truth answers \citep{kim2023computer, shinn2023reflexion}, human feedback \citep{chen2024learning}, or stronger models \citep{zhang2024small}.

\paragraph{Fine-tuning ({\normalfont \S\ref{sec:fine-tuning}}).}
Models fine-tuned for self-correction are another source of feedback, which are trained via supervised fine-tuning \citep{welleck2023generating, selfee2023, first2023baldur, paul-etal-2024-refiner, han2024small, havrilla2024glore} or reinforcement learning \citep{le2022coderl, akyurek-etal-2023-rl4f}.

\subsection{Tasks}
Self-correction has been studied in various tasks, including {\bf Reasoning:}~arithmetic reasoning \citep{madaan2023selfrefine, nathani-etal-2023-maf, gou2024critic}, code generation \citep{jiang2023selfevolve, charalambous2023new, gou2024critic, chen2024teaching, olausson2024is}, proof generation \citep{first2023baldur}, logical reasoning \citep{pan-etal-2023-logic}, {\bf Knowledge:}~closed-book QA \citep{shinn2023reflexion, gao-etal-2023-rarr, jiang-etal-2023-active, gou2024critic}, {\bf Context-based Generation:}~dialogue generation \citep{madaan2023selfrefine, peng2023check}, text summarization \citep{saunders2022selfcritiquing}, {\bf Open-ended Generation:} conditional text generation \citep{selfee2023, schick2023peer}, story generation \citep{yang-etal-2022-re3}, detoxification \citep{timo2021selfdiagnosis, bai2022constitutional, gou2024critic, phute2024llm}, {\bf Others:}~machine translation \citep{chen2023iterative, raunak-etal-2023-leveraging, ki2024guiding}, information retrieval \citep{gero2023selfverification}, vision language tasks \citep{yin2023woodpecker, ge-etal-2023-wrong, zhou2024analyzing, lee2023volcano, huang2024lvlms, liu2024intrinsicselfcorrectioncapabilityllms}, and prompt optimization \citep{pryzant-etal-2023-automatic, mehrabi2023flirt, yang2024large}.

\subsection{Differences from Related Approaches}

In this work, we define self-consistency \citep{wang2023selfconsistency} or generate-and-rank \citep{shen-etal-2021-generate-rank, weng-etal-2023-large} to be different from self-correction because these approaches do not refine responses and assume that LLMs generate correct answers with a reasonable probability.
We discuss these methods in Section~\ref{sec:comparable-baselines} as strong baselines that should be compared with self-correction.

\section{Research Questions} \label{sec:target-claims}

We find that prior studies often do not define their research questions in detail and fail to use appropriate self-correction frameworks in their experiments. We propose a new approach to classify research questions and frameworks in self-correction.

\subsection{RQs in Self-Correction Research} \label{sec:claims}

Prior studies often simply state their research questions as {\it whether LLMs can self-correct their mistakes} \citep[e.g.,][]{kim2023computer, madaan2023selfrefine}. However, we claim that \ul{research questions in self-correction research should be defined in more detail}. We identify the following research questions implicitly targeted in prior studies, as in Table~\ref{tab:claims-in-prior-work}.

\begin{itemize}
    \item \mbox{[\rqone]}~\phantomsection\label{claim-a}Can LLMs self-correct their best-possible initial responses {\it based solely on the inherent capabilities?}~(\S\ref{sec:prompting-self-correction})
    \item \mbox{[\rqtwo]}~\phantomsection\label{claim-b}Can LLMs self-correct their best-possible initial responses {\it assisted by external information?}~(\S\ref{sec:external-info})
    \item \mbox{[\rqthree]}~\phantomsection\label{claim-c}Are the final outputs from self-correction {\it better than other methods?}~(\S\ref{sec:comparable-baselines})
\end{itemize}

\noindent
We define the {\it best-possible initial responses} as \ul{initial responses generated with best effort, using information that self-correction modules can access}, such as external tools, knowledge, or fine-tuning.

\paragraph{Requirements for Verifying RQs.} Experiments for verifying these research questions need to satisfy different requirements, as shown in Table~\ref{tab:claims-requirements}. {\bf External Information:}~\rqone{} needs to be evaluated on frameworks that refine responses using the same model without additional information. \rqtwo{} and \rqthree{} can be evaluated on frameworks that use external information. {\bf Initial Responses:}~\rqone{} and \rqtwo{} need to be evaluated on frameworks that use the {\it best-possible initial responses}. \rqthree{} is about the final performance, so it is not necessary to start from strong initial responses. {\bf Evaluation:}~\rqone{} and \rqtwo{} only require to show that self-correction improves performance from the initial responses. \rqthree{} requires comparison with strong baselines~(\S\ref{sec:comparable-baselines}).

\paragraph{Confusion in Prior Work.}
Some prior studies implicitly target different research questions in a single work without clearly distinguishing them. As in Table~\ref{tab:claims-in-prior-work}, \citet{kim2023computer} target \rqone{} for arithmetic reasoning by comparing self-corrected responses only with initial responses, but they target \rqthree{} for MiniWoB++ by comparing self-correction with baseline methods. Similarly, \citet{gou2024critic} target \rqtwo{} for arithmetic reasoning but target \rqthree{} for detoxification.

\subsection{Frameworks for Verifying RQs} \label{sec:framework-categories}
Prior work often categorizes self-correction frameworks based on approaches for generating feedback~(\S\ref{sec:overview}). However, we point out that we also need to categorize them by \ul{the quality of initial responses} because the frameworks we need to use for verifying different research questions vary by whether they use the best-possible initial responses~(\S\ref{sec:claims}).

We propose categories of (same-model) self-correction that correspond to different research questions~(\S\ref{sec:claims}), as shown in Figure~\ref{fig:categories}. Specifically, we propose to categorize the self-correction frameworks as follows.

\begin{itemize}\vspace{2pt}
\item Realistic: Can be used in real-world applications.
\begin{itemize}
    \item Fair: Using best-possible initial responses
    \item Unfair: Using sub-optimal initial responses
\end{itemize}
\item Unrealistic: Using information that is not accessible in real-world applications.
\end{itemize}\vspace{2pt}

\noindent
In this work, we focus on categorizing self-correction frameworks that do not involve multiple language models with different architectures. Cross-model correction uses different models for initial response generation and self-correction, so it is unsuitable for evaluating whether LLMs can improve their own initial responses [\rqone, \rqtwo]. However, it can be used to evaluate [\rqthree] whether the final responses from self-correction are better than other methods.

\paragraph{Realistic vs. Unrealistic.} Some prior studies propose unrealistic self-correction, which cannot be implemented in real-world applications, by using oracle information such as ground-truth answers \citep{kim2023computer, shinn2023reflexion}. These methods cannot be used to verify any research questions.

\paragraph{Fair vs. Unfair.} Realistic frameworks can be categorized by whether they use the best-possible initial responses.
{\bf Fair self-correction} represents frameworks that refine the best-possible initial responses. (1)~{\it Intrinsic self-correction} \citep{huang2024large} uses the same model and information for initial response generation and self-correction. Intrinsic self-correction can be used to assess [\rqone] whether LLMs can self-correct based solely on their inherent capabilities. (2)~{\it Fair-asymmetric self-correction} uses additional information for self-correction, but also uses information to improve initial response generation as much as possible. For example, self-correction with code interpreters \citep{chen2024teaching, gou2024critic} is not intrinsic but fair because we cannot easily use code interpreters to directly improve the initial response generation. Fair-asymmetric self-correction can be used to evaluate [\rqtwo] whether LLMs can self-correct the best-possible initial responses using external information.
{\bf Unfair self-correction} (or {\it unfair-asymmetric self-correction}) represents frameworks that are practical but do not use the best-possible initial responses. For example, methods that use search engines only for self-correction \citep{gao-etal-2023-rarr, yu2023improving} are unfair because they can use search engines to directly improve the initial response generation. Unfair self-correction can evaluate [\rqthree] whether the final responses from self-correction outperform other methods but cannot evaluate [\rqtwo] whether self-correction can improve the best-possible initial responses.

\begin{table*}[t]
\renewcommand{\arraystretch}{.5}
\fontsize{7pt}{7pt}\selectfont
\centering
\begin{tabular}{M{.1\textwidth}M{.135\textwidth}M{.13\textwidth}M{.13\textwidth}M{.38\textwidth}}
\toprule
Paper & Task & Using Oracle Info for~Feedback & Weak Prompt for Initial Responses & Comments  \\
\midrule
\phantom{----}RCI\phantom{----} \citeyearpar[\S3.1]{kim2023computer} & Computer Tasks
& \phantom{---------}\checkmark{}\phantom{---------} stop~condition  & --         
& Using ground-truth answers and do not update correct responses, which unfairly ignores false-positive correction \\
\midrule
Reflexion \mbox{\citeyearpar[\S4.2]{shinn2023reflexion}} & HotpotQA (Context)
& \phantom{---------}\checkmark{}\phantom{---------} feedback        & --         
& Feedback is the exact match between the responses and ground-truth answers \\
\midrule
CAI~Revisions \citeyearpar{bai2022constitutional} & Detoxification
& --          & \checkmark
& Initial generation is not prompted to remove harmful outputs \\
\midrule
Self-Refine \citeyearpar{madaan2023selfrefine} & Math, Coding, Dialogue
& --          & \checkmark 
& Unfairly weak or wrong instructions or few-shot demonstrations for initial response generation \\
\bottomrule
\end{tabular}
\caption{Unfair settings in prior studies of self-correction with prompting, over-evaluating self-correction.} \label{tab:unfair-prior-work}
\end{table*}

\section{Self-Correction with Prompting} \label{sec:prompting-self-correction}

\begin{quote}
[\rqone{}] Can LLMs self-correct their best-possible initial responses {\it based solely on the inherent capabilities?}
\end{quote}

Several studies propose {\it intrinsic self-correction} methods, which self-correct responses from LLMs by prompting themselves to generate feedback and refine the responses. \citet{bai2022constitutional} propose self-correcting harmful responses from LLMs by prompting themselves. Self-Refine \citep{madaan2023selfrefine} and RCI Prompting \citep{kim2023computer} iteratively prompt LLMs to self-correct their own responses in tasks such as arithmetic reasoning.

\paragraph{Negative Results.} However, recent studies report that intrinsic self-correction does not improve or even degrade the performance in tasks such as arithmetic reasoning, closed-book QA \citep{huang2024large, gou2024critic}, code generation \citep{gou2024critic, olausson2024is}, plan generation \citep{valmeekam2023investigating}, and graph coloring \citep{stechly2023gpt}.
Several studies claim that a bottleneck is in the feedback generation, and it is difficult to generate reliable feedback on their responses only by prompting themselves \citep{gou2024critic, huang2024large, olausson2024is, chen-etal-2024-tree}.

\paragraph{Unrealistic or Unfair Settings.} The conflicting positive and negative results motivate us to analyze when LLMs can self-correct {\it only by prompting themselves}. Specifically, we assess whether prior studies satisfy the requirements to verify that [\rqone]~LLMs can self-correct their responses based solely on their inherent capabilities.
As in Table~\ref{tab:unfair-prior-work}, we find that many studies use either oracle information in the self-correction processes (unrealistic frameworks) or weak prompts that can be easily improved for generating initial responses (unfair settings), which over-evaluate self-correction. Consequently, we conclude that \ul{no major work shows successful self-correction of responses from LLMs using feedback generated by prompting themselves under fair settings in general tasks}.
{\bf Oracle Information:}~RCI Prompting \citep{kim2023computer} uses ground-truth answers and does not apply self-correction when the initial responses are correct, which unfairly ignores mistakes caused by updating correct responses incorrectly. Reflexion \citep{shinn2023reflexion} generates feedback by using an exact match between the generated and ground-truth answers, which cannot be accessed in real-world applications.
{\bf Weak Initial Responses:}~Detoxifying harmful responses is a popular task in self-correction research, but prior studies often study in situations where initial response generation is not instructed to generate harmless responses \citep{bai2022constitutional, wang2024a}. Although detecting harmful contents using LLMs is a reasonable research topic, this setting is not the self-correction from best-possible initial responses, since we can improve the initial response generation process by instructing not to generate harmful responses. As more obvious weak prompts, Self-Refine \citep{madaan2023selfrefine} uses instructions or few-shot examples that do not correctly correspond to the target task only for initial response generation (e.g., providing wrong target labels in few-shot examples), while using appropriate instructions for self-correction, as shown in Table~\ref{tab:self-refine-dialogue-response} and \ref{tab:self-refine-sentiment-reversal}. These settings evaluate improvement from weak initial responses, which over-evaluate the improvement by self-correction.

\paragraph{Tasks in which Self-Correction is Exceptionally Effective.}
Although our analysis of prior studies shows that intrinsic self-correction is difficult in general, some tasks have properties that make feedback generation easy and enable intrinsic self-correction. For example, CoVe \citep{dhuliawala2023chainofverification} is an intrinsic self-correction method for tasks of generating multiple answers, such as {\it Name some politicians who were born in NY, New York.} Generated responses include multiple answers, but the feedback generation can be decomposed into easier sub-tasks of verifying each answer. Tasks with {\bf decomposable responses} are one of the few groups of tasks for which verification is clearly easier than generation, which enables intrinsic self-correction. However, many real-world tasks do not satisfy this property.

\begin{table*}[t]
    \newcommand{\tworows}[1]{\multirow{2.7}{\linewidth}{\centering #1}}
    \fontsize{7pt}{7pt}\selectfont
    \centering
    \begin{tabular}{M{.2\textwidth}M{.25\textwidth}
    M{.2\textwidth}M{.22\textwidth}}
        \toprule
        \tworows{Paper} & \tworows{Main Task} & \multicolumn{2}{c}{External Tools or Knowledge} \\
        \cmidrule(l{2pt}r{2pt}){3-4}
        && For Initial Response Generation & For Feedback Generation \\
        \midrule
        Reflexion \citeyearpar[\S4.1, 4.3]{shinn2023reflexion}  &
        Games, Coding &
        -- & Game Envs, Code~Interpreter \\
        CRITIC \citeyearpar{gou2024critic} &
        GSM8k, SVAMP &
        -- & Python interpreter \\
        Self-Debug \citeyearpar{chen2024teaching} &
        Text-to-Code &
        -- & Code Interpreter \\
        \midrule
        CRITIC \citeyearpar{gou2024critic} &
        HotpotQA &
        Web Search & Web Search \\
        FLARE \citeyearpar{jiang-etal-2023-active} &
        2WikiMultihopQA, StrategyQA, ASQA &
        Web Search & Web Search \\
        \midrule
        RARR \citeyearpar{gao-etal-2023-rarr} &
        NQ, SQA, QReCC &
        -- & Web Search \\
        ReFeed \citeyearpar{yu2023improving} &
        NQ, TriviaQA, HotpotQA &
        -- & Wikipedia \\
        \bottomrule
    \end{tabular}
    \caption{Self-correction with external tools or knowledge (with in-context learning).} \label{tab:external-feedback}
\end{table*}

\section{Self-Correction with External Information} \label{sec:external-info}

\begin{quote}
[\rqtwo{}] Can LLMs self-correct their best-possible initial responses {\it assisted by external information?}
\end{quote}

This section analyzes self-correction frameworks that make use of external tools, external knowledge, and fine-tuning.

\subsection{Self-Correction with External Tools or Knowledge} \label{sec:external-feedback}

Given the observation that feedback generation is a bottleneck of self-correction (\S\ref{sec:prompting-self-correction}), improving feedback using external tools or knowledge is a promising direction. External tools used for self-correction include code interpreters for code generation tasks \citep{chen2024teaching, gou2024critic} and symbolic reasoners for logical reasoning tasks \citep{pan-etal-2023-logic}. A popular source of knowledge is search engines, which are often used with queries generated from initial responses to retrieve information for validating their correctness \citep{gao-etal-2023-rarr, jiang-etal-2023-active}. These prior studies widely agree that self-correction can improve LLM responses when reliable external tools or knowledge suitable for improving feedback are available.

\paragraph{Unfair self-correction with external information.}
Although using external tools or knowledge is known to be effective in self-correction, we raise caution that the way of using external tools or knowledge influences the research questions we can verify (\S\ref{sec:claims}). As shown in Table~\ref{tab:external-feedback}, some prior studies \citep{gao-etal-2023-rarr, yu2023improving, zhao-etal-2023-verify} use external knowledge only for self-correction, while they can also directly use external knowledge to improve the initial response generation process. For example, RARR \citep{gao-etal-2023-rarr} uses external knowledge to detect mistakes in initial responses, while it does not use any external knowledge when generating initial responses. \ul{These methods are reasonable when only focusing on [\rqthree]~the performance of final responses, but it is not fair to use them for evaluating [\rqtwo]~whether self-correction can improve from the best-possible initial responses}. In contrast, using code interpreters for self-correction \citep{gou2024critic, chen2024teaching} can be regarded as using best-possible initial responses because there is no easy way to improve the initial response generation directly.

\paragraph{Verifiable Tasks.}
Some tasks have a property that allows the correctness of the responses to be verified easily, even without external information. For example, the constrained generation task evaluated in Self-Refine \citep{madaan2023selfrefine} is a task to generate a sentence that includes five specified words. We can easily evaluate the correctness by checking whether the five words are included in the generated sentence. Tree-of-thought \citep{yao2023tree-of-thought} is a generate-and-rank method for verifiable tasks,\footnote{Tree-of-thought is a generate-and-rank method and not a self-correction method in our definition.} such as Game of 24, the task to obtain 24 using basic arithmetic operations ($+, -, \times, \div$) and provided four integers. For Game of 24, we can easily verify the answer by checking whether the generated answer is 24. We consider self-correction to work well in these tasks because they are in the same situations as using strong external tools or the oracle information to generate feedback.

\begin{table*}[t]
    \newcommand{\tworows}[1]{\multirow{2.7}{\linewidth}{\centering #1}}
    \renewcommand{\arraystretch}{.5}
    \setlength{\tabcolsep}{3pt}
    \fontsize{7pt}{7pt}\selectfont
    \centering
    \begin{tabular}{M{.1\linewidth}M{.1\linewidth}
    M{.055\linewidth}M{.06\linewidth}
    M{.075\linewidth}M{.09\linewidth}
    M{.075\linewidth}M{.09\linewidth}M{.05\linewidth}
    M{.075\linewidth}M{.09\linewidth}}
        \toprule
        \tworows{Paper} & \tworows{Main Task} & \tworows{Cross-Model} & \tworows{SFT Tasks} &
        \multicolumn{2}{c}{Initial Responses} & \multicolumn{3}{c}{Feedback Generation} & \multicolumn{2}{c}{Refinement} \\
        \cmidrule(l{2pt}r{2pt}){5-6} \cmidrule(l{2pt}r{2pt}){7-9} \cmidrule(l{2pt}r{2pt}){10-11}
        &&&& Model & SFT Target & Model & SFT Target & Size & Model & SFT Target \\
        \midrule
        \phantom{--}SelFee\phantom{--} \citeyearpar{selfee2023} &
        MT-Bench & -- & General Tasks &
        \textcolor{Orange}{Llama (7B,13B)} & ChatGPT Responses &
        \textcolor{Orange}{Llama (7B,13B)} & ChatGPT Feedback & 178K &
        \textcolor{Orange}{Llama (7B,13B)} & ChatGPT Refinement \\
        \midrule
        \phantom{--}Volcano\phantom{--} \citeyearpar{lee2023volcano} &
        Visual Reasoning & -- & General Tasks &
        \textcolor{Orange}{LLaVA (7B,~13B)} & GPT-3.5-T, Human &
        \textcolor{Orange}{LLaVA (7B,~13B)} & GPT-3.5-T Feedback & 274K & 
        \textcolor{Orange}{LLaVA (7B,~13B)} & Reference Answers \\
        \midrule
        Self-Critique \citeyearpar{saunders2022selfcritiquing} &
        Topic-based Summarization & -- & Target Task &
        \textcolor{Orange}{Instruct GPT} & Human Summaries &
        \textcolor{Orange}{Instruct GPT} & Human Feedback & 100K &
        \textcolor{Orange}{Instruct GPT} & Human Refinement \\
        \midrule
        REFINER \citeyearpar{paul-etal-2024-refiner} &
        Math, Logic, Moral Stories & \checkmark & Target Task &
        \textcolor{Purple}{GPT-3.5} & -- &
        \textcolor{Green}{T5-base} & Synthetic Data & 20K - 30K &
        \textcolor{Purple}{GPT-3.5} & -- \\
        \midrule
        Self-Edit \citeyearpar{zhang-etal-2023-self} &
        Code Generation & \checkmark & Target Task &
        \textcolor{Purple}{GPT-3} & -- &
        \multicolumn{3}{c}{(Code Executor and Test Cases)} &
        \textcolor{Green}{PyCodeGPT 110M} & Reference Code \\
        \bottomrule
    \end{tabular}
    \caption{Self-correction with supervised fine-tuning. Most methods require large training datasets. \mbox{``{--}''}~on the ``SFT Target'' columns represents no fine-tuning.} \label{tab:finetuning-strategies}
\end{table*}

\begin{table*}[t]
    \scriptsize
    \centering
    \renewcommand{\arraystretch}{1.2}
    \setlength{\tabcolsep}{3.5pt}
    \begin{tabular}{M{.03\linewidth}M{.03\linewidth}M{.03\linewidth}L{.68\linewidth}cM{.1\linewidth}}
    \toprule
        \rqone & \rqtwo & \rqthree & \multicolumn{1}{c}{Requirements for Verifying the Target RQs} & &  \\
    \midrule
        \checkmark & \checkmark & \checkmark &
        Clearly stating the target RQ and the category of self-correction framework discussed.
        & (\S\ref{sec:framework-categories}) & \textcolor{red}{Required} \\
        \checkmark & \checkmark & \checkmark &
        Not using oracle information, such as ground-truth answers.
        & (\S\ref{sec:prompting-self-correction}) & \textcolor{red}{Required} \\
        \checkmark & \checkmark & \checkmark &
        {\it When using fine-tuning}, reporting the detailed settings, including the number of annotations and computational cost required to achieve the reported performance.
        & (\S\ref{sec:fine-tuning}) & \textcolor{red}{Required} \\
        \checkmark & \checkmark & \checkmark &
        Evaluating the quality of feedback directly (e.g., error detection accuracy).
        & (\S\ref{sec:conclusions-of-analysis}) & \textcolor{blue}{Recommended} \\
    \midrule
        \checkmark & \checkmark &            &
        Using sufficiently strong prompts for generating initial responses.
        & (\S\ref{sec:prompting-self-correction}) & \textcolor{red}{Required} \\
    \midrule
        \checkmark &            &            &
        Using intrinsic self-correction.
        & (\S\ref{sec:framework-categories}) & \textcolor{red}{Required} \\
    \midrule
        &&& \multicolumn{1}{c}{\it When using external tools or knowledge,} \\
                   & \checkmark &            &
        Using external tools or knowledge to improve initial response generation as much as possible.
        & (\S\ref{sec:external-feedback}) & \textcolor{red}{Required} \\
    \midrule
        &&& \multicolumn{1}{c}{\it When using fine-tuning for self-correction,} \\
                   & \checkmark &            &
        Fine-tuning initial response generators as well, as much as possible.
        & (\S\ref{sec:fine-tuning}) & \textcolor{red}{Required} \\
                   & \checkmark &            &
        Evaluating the minimum required size of training data that enables self-correction.
        & (\S\ref{sec:fine-tuning}) & \textcolor{blue}{Recommended} \\
                   & \checkmark &            &
        Evaluating cross-model correction setting that refines mistakes in responses from stronger LLMs.
        & (\S\ref{sec:framework-categories}) & \textcolor{blue}{Recommended} \\
    \midrule
                   &            & \checkmark &
        Comparing with strong baselines using comparable computational cost.
        & (\S\ref{sec:comparable-baselines}) & \textcolor{red}{Required} \\
    \bottomrule
    \end{tabular}
    \caption{Checklist for self-correction research for different target research questions.}
    \label{tab:checklist}
\vskip 1.5em
    \scriptsize
    \centering
    \renewcommand{\arraystretch}{1.4}
    \setlength{\tabcolsep}{5pt}
    \begin{tabular}{L{.8\linewidth}cM{.1\linewidth}}
    \toprule
        Clearly stating the RQ that is refuted by the reported results and the category of the framework discussed.
        & (\S\ref{sec:framework-categories}) & \textcolor{red}{Required} \\
        Using strong prompts for self-correction (e.g., state-of-the-art reference-free metrics).
        & (\S\ref{sec:future-directions}) & \textcolor{red}{Required} \\
        {\it When not using external tools or knowledge available in real-world applications}, explicitly reporting that the evaluation is done under weak conditions.
        & (\S\ref{sec:external-feedback}) & \textcolor{red}{Required} \\
        Evaluating with external tools or knowledge available in real-world applications.
        & (\S\ref{sec:external-feedback}) & \textcolor{blue}{Recommended} \\
    \bottomrule
    \end{tabular}
    \caption{Checklist for reporting negative results of self-correction.}
    \label{tab:checklist_negative_results}
\end{table*}

\subsection{Self-Correction with Fine-tuning} \label{sec:fine-tuning}

Prior work shows that fine-tuning LLMs for generating feedback or refining responses improves the self-correction capability. A common approach fine-tunes feedback models to generate reference feedback given initial responses and fine-tunes refinement models to generate reference answers given the initial responses and reference feedback \citep{selfee2023, lee2023volcano, saunders2022selfcritiquing}.
{\bf Frameworks:}~The first approach fine-tunes {\it the same model} to correct its own responses. In this approach, most methods fine-tune models for all stages: initial responses, feedback, and refinement \citep{saunders2022selfcritiquing, selfee2023, lee2023volcano}.
Another approach corrects responses from larger models using {\it smaller fine-tuned models}. This cross-model correction approach often instructs the larger models to refine their own responses using feedback from the smaller fine-tuned models \citep{yang-etal-2022-re3, welleck2023generating, akyurek-etal-2023-rl4f, paul-etal-2024-refiner}, which can be viewed as using the small fine-tuned models as external tools.
{\bf Training Strategies:}~A popular approach is supervised fine-tuning, which fine-tunes self-correction modules on human-annotated feedback \citep{saunders2022selfcritiquing}, feedback from stronger models \citep{selfee2023}, or synthetic negative responses \citep{paul-etal-2024-refiner}. As other approaches, to avoid the cost of collecting human feedback, self-corrective learning \citep{welleck2023generating} selects model-generated feedback that successfully refines responses as training data, GLoRe \citep{havrilla2024glore} constructs a synthetic refinement dataset using model-generated feedback, and RL4L \citep{akyurek-etal-2023-rl4f} uses reinforcement-learning.
{\bf External Tools:}~Some works fine-tune models to refine responses given feedback from external tools.
Self-Edit \citep{zhang-etal-2023-self} uses the results on test cases evaluated by code executors for code generation, and Baldur \citep{first2023baldur} uses proof assistants for improving proof generation.

\paragraph{Large Training Data for SFT of Feedback.}
As shown in Table~\ref{tab:finetuning-strategies}, many methods with supervised fine-tuning for feedback generation rely on training data with more than 100K instances. These studies often use feedback generated by stronger models to simulate human annotation, but this approach requires large-scale human annotations to be implemented on state-of-the-art models. We expect future research to explore approaches that do not require large-scale human annotations (\S\ref{sec:future-directions}).

\paragraph{Unfair Fine-tuning.} Some studies \citep{welleck2023generating} apply stronger fine-tuning for self-correction models than initial response generation models, which do not use best-possible initial responses in the available resources (\S\ref{sec:framework-categories}). This approach can be used to evaluate [\rqthree] the performance of the final responses to compare with other methods but cannot be used to evaluate [\rqtwo] the improvement from best-possible initial responses.

\section{Strong Baselines} \label{sec:comparable-baselines}

\begin{quote}
[\rqthree{}] Are the final outputs from self-correction {\it better than other methods?}
\end{quote}

Self-correction involves multiple LLM calls for generating feedback and refinement. Therefore, to claim that [\rqthree{}]~the performance of the final outputs from self-correction frameworks is better than other approaches, it should be compared with sufficiently strong baselines, possibly relying on additional LLM calls or computational cost. Many self-correction studies do not compare their methods with strong baselines, although some studies pointed out this issue and compare self-correction with self-consistency \citep{gou2024critic, huang2024large} or pass@k in code generation \citep{zhang-etal-2023-self, olausson2024is}. We encourage future research to compare self-correction with strong baselines, including self-consistency and generate-and-rank, to further explore \rqthree{}.

\paragraph{Self-Consistency} \citep{wang2023selfconsistency} is an approach that generates multiple responses for the same input and takes the majority vote of the final answers in reasoning tasks. The idea of selecting good responses using the consistency between multiple responses from the same model has also been extended to other tasks such as text generation \citep{manakul-etal-2023-selfcheckgpt, elaraby2023halo, chen2023universal} and code generation \citep{shi-etal-2022-natural}.

\paragraph{Generate-and-Rank}
is an approach that generates multiple responses and selects the best response using verifiers.
{\bf Post-hoc} approach ranks responses using self-evaluation \citep{weng-etal-2023-large, zhang23coder}, confidence \citep{manakul-etal-2023-selfcheckgpt}, fine-tuned verifiers \citep{cobbe2021training, shen-etal-2021-generate-rank, lightman2024lets}, or verifiers with external tools \citep{shi-etal-2022-natural, chen2023codet, ni23lever}.
{\bf Feedback-guided decoding} generates multiple responses and selects the best response for each reasoning step using generation probability \citep{hao-etal-2023-reasoning, tyen2024llms}, prompted self-evaluation \citep{jung-etal-2022-maieutic, creswell2022faithful, xie2023selfevaluation, yao2023tree-of-thought, miao2024selfcheck}, or fine-tuned verifiers \citep{uesato2022solving, tafjord-etal-2022-entailer, yang-etal-2022-generating, asai2024selfrag}.

\section{Summary of Our Analysis} \label{sec:conclusions-of-analysis}

\paragraph{Bottleneck is in Feedback Generation.}
Prior studies widely agree that LLMs can {\it refine} their responses given reliable feedback (\S\ref{sec:external-info}). However, generating reliable {\it feedback} on their own responses is still observed to be challenging for LLMs without using additional information (\S\ref{sec:prompting-self-correction}). In other words, for the current LLMs, the hypothesis that {\it recognizing errors is easier than avoiding them} \citep{saunders2022selfcritiquing} is only true for certain tasks whose verification is exceptionally easy, according to our analysis of the experiments in prior studies. We recommend that self-correction research analyze the quality of generated feedback in more detail, not only evaluate the downstream performance of the refined responses.

\paragraph{Tasks Suitable for Self-Correction.}
Our analysis identifies the properties of tasks that are suitable for self-correction under different conditions.

\begin{itemize}
    \item Intrinsic Self-Correction (\S\ref{sec:prompting-self-correction})
    \begin{itemize}
        \item Tasks whose verification tasks are much easier than the original tasks (e.g., tasks whose responses are decomposable)
    \end{itemize}
    \item Self-Correction with External Information (\S\ref{sec:external-feedback})
    \begin{itemize}
        \item Tasks for which external tools that provide reliable feedback exist (e.g., code generation)
        \item Tasks for which responses can be utilized to obtain useful information that is difficult to obtain before generating initial responses (e.g.,~generate queries from responses to retrieve documents for verifying information)
    \end{itemize}
    \item Self-Correction with Fine-tuning (\S\ref{sec:fine-tuning})
    \begin{itemize}
        \item Self-correction works in many tasks when large training data for feedback generation is available
        \item Tasks that can use reinforcement learning or self-corrective learning \citep{welleck2023generating}, i.e.,~tasks whose responses can be easily evaluated given ground-truth answers
    \end{itemize}
\end{itemize}

\section{Checklist for Self-Correction Research} \label{sec:checklist}

Our analysis shows that many studies do not clearly define their research questions and fail to conduct appropriate experiments (\S\ref{sec:claims}, \ref{sec:prompting-self-correction}). To tackle these issues, we provide a checklist for self-correction research that provides requirements for designing appropriate experiments for verifying target RQs and recommended experiments for comprehensive analysis. Table~\ref{tab:checklist} provides a checklist for verifying different RQs identified in Section~\ref{sec:claims}. Table~\ref{tab:checklist_negative_results} provides a checklist for reporting negative results.

\section{Differences from Other Survey} \label{sec:other-survey}

\citet{pan2024automatically} provide a comprehensive survey on broad topics related to self-correction, including training strategies. Our work specifically focuses on (inference-time) self-correction and provides a more detailed and critical analysis of prior work.
\citet{huang2024large} provide an analysis of problems in the evaluation settings of self-correction research, which motivates our work. They focus on analyzing a few papers on intrinsic self-correction in reasoning tasks. We provide a more comprehensive analysis of self-correction with in-context learning, external tools, and fine-tuning.

\section{Related Work of Self-Correction} \label{sec:related-work}

\paragraph{Self-Detection} of mistakes in LLM responses using LLMs (possibly with external information) has been studied in various domains, including misinformation detection \citep{zhang2023language, chern2023factool, chen2024can, mishra2024finegrained}, context-faithfulness \citep{wang-etal-2020-asking, durmus-etal-2020-feqa, scialom-etal-2021-questeval}, harmful content detection \citep{rauh2022characteristics}, and bias detection \citep{blodgett-etal-2020-language, feng-etal-2023-pretraining}. However, recent studies \citep{tyen2024llms, kamoi2024evaluating} show that even strong LLMs often cannot detect their own mistakes in various tasks.

\ifreview\else
\paragraph{Editing Human-Written Text} by using language models has been studied in various domains, including information update \citep{Shah_Schuster_Barzilay_2020, iv-etal-2022-fruit, schick2023peer}, grammatical error correction \citep{ng-etal-2014-conll, lichtarge-etal-2019-corpora}, factual error correction \citep{cao-etal-2020-factual, thorne-vlachos-2021-evidence}, and code repair \citep{gupta2017deepfix, mesbah2019deepdelta, bader2019getafix, chen2021sequencer, yasunaga2020graph, yasunaga2021break-it-fix-it}.
\fi

\paragraph{Self-Training} or self-improvement is an approach to train models using their own responses. Some studies use self-evaluation or self-correction for creating training data \citep{bai2022constitutional, gulcehre2023reinforced} or use self-evaluation as training signals \citep{pang2024language}. Another approach improves the reasoning of LLMs using LLM-generated reasoning by selecting high-quality outputs using ground-truth final answers \citep{zelikman2022star} or self-consistency \citep{huang-etal-2023-large}. As another direction, \citet{yu2022generating} use sentences generated by LLMs with high confidence for training classifiers.

\section{Future Directions} \label{sec:future-directions}

\paragraph{Improving Feedback.}
Prior studies indicate that it is difficult for LLMs to generate feedback on their own responses with in-context learning (\S\ref{sec:prompting-self-correction}, \ref{sec:conclusions-of-analysis}). However, most studies in intrinsic self-correction \citep{madaan2023selfrefine, huang2024large} use simple prompts for generating feedback, and there is room for improvement. A possible direction to improve feedback is to apply (reference-free and point-wise) {\bf LLM-based evaluation metrics}. Recent approaches for improving the model-based evaluation include using human-written evaluation criteria \citep{chiang-lee-2023-large, liu-etal-2023-g} and decomposing responses \citep{saha2023branchsolvemerge, min-etal-2023-factscore}. As another direction, recent studies in self-correction propose frameworks using the {\bf confidence} in their responses, estimated by generation probabilities \citep{varshney2023stitch, jiang-etal-2023-active}, prompting \citep{li2024confidencematters}, or generating new questions from their answers to evaluate logical consistency \citep{jung-etal-2022-maieutic, tafjord-etal-2022-entailer, wu2024large}.

\paragraph{Unexplored Tasks.}
The difficulty of self-evaluation differs from task to task (\S\ref{sec:prompting-self-correction}), while many studies assume that verification is consistently easier than generation. We expect that there are unexplored tasks in which intrinsic self-correction works well, although self-correction research mostly focuses on reasoning tasks such as math reasoning and coding \citep{madaan2023selfrefine, gou2024critic, huang2024large}.
For example, LLM-based evaluation is often studied in open-ended text generation, such as dialogue generation and text summarization \citep{Fu2023GPTScore, liu-etal-2023-g}, suggesting that reasonable model-based feedback is available for these tasks.

\paragraph{Fine-tuning on Small Training Data.}
Fine-tuning of feedback generation often relies on large training data, which requires large-scale human annotations (\S\ref{sec:fine-tuning}). We expect future work to explore self-correction with smaller training data.
Although reinforcement learning \citep{akyurek-etal-2023-rl4f} or self-corrective learning \citep{welleck2023generating} do not require human feedback, they require reasonable reward functions for evaluating LLM responses, which are not available in many tasks. For example, RL4F \citep{akyurek-etal-2023-rl4f} uses ROUGE as a reward function for text summarization and action planning, which is sub-optimal.

\paragraph{Pre-training for Improving Self-Correction.} Prior studies show that large-scale fine-tuning on reference feedback improves the self-correction capability of LLMs (\S\ref{sec:fine-tuning}). This observation suggests that the current approach or datasets for pre-training LLMs are insufficient to make LLMs acquire self-correction capability. We expect future work to explore pre-training strategies to improve the intrinsic self-correction capability of LLMs.

\section{Emerging Trends and Recent Developments} \label{sec:recent-papers}

This survey, originally published in June 2024, primarily focuses on papers published before May 2024. However, to provide a broader perspective, this section briefly highlights emerging trends and recent advancements from June 2024 onward.

A recent trend of self-correction involves employing reinforcement learning \citep{kumar2024training, qu2024recursive}. Specifically, OpenAI has published o1 \citep{openai2024o1}, a model for reasoning tasks trained with reinforcement learning to explore different strategies, recognize their own mistakes, and refine their thinking process. OpenAI~o1 has been reported to outperform state-of-the-art LLMs in various reasoning tasks, including Math Olympiad, PhD-level academic problems, competitive programming, and Kaggle \citep{chan2024mlebench}.

\section{Conclusion}

We provide a critical survey of self-correction to identify in which conditions LLMs can self-correct their mistakes. Our analysis reveals that many studies fail to define their research questions clearly or design experiments appropriately. To tackle these issues, we categorize research questions and frameworks in self-correction research and provide a checklist for conducting appropriate experiments.

\ifreview\else
\section*{Acknowledgments}

This work was supported by a Cisco Research Grant. We appreciate valuable suggestions from the action editor and anonymous reviewers.

\fi

\bibliography{custom}
\bibliographystyle{acl_natbib}

\appendix

\begin{table*}[t]
    \centering
    \fontsize{8pt}{8pt}\selectfont
    \begin{tabular}{L{.47\linewidth}L{.47\linewidth}}
    \toprule
       \multicolumn{1}{c}{Initial Response Prompt} & \multicolumn{1}{c}{Feedback Prompt} \\
    \midrule
Provided a dialogue between two speakers, generate a response that is coherent with the dialogue history. {\bf Desired traits for responses} are: 1) Relevant - The response addresses the context, 2) Informative - The response provides some information, 3) Interesting - The response is \textbf{\textcolor{red}{not interesting}}, 4) Consistent - The response is consistent with the rest of the conversation in terms of tone and topic, 5) Helpful - The response is helpful in providing any information or suggesting any actions, 6) Engaging - The response is \textbf{\textcolor{red}{not very engaging and does not encourage further conversation}}, 7) Specific - The response contains pecific content, 9) User understanding - The response demonstrates an understanding of the user's input and state of mind, and 10) Fluent. Response should begin with - Response:

\vskip .5em
\textcolor{teal}{[3 examples omitted]}
        &
We want to iteratively improve the provided responses. To help improve, scores for each response on desired traits are provided: 1) Relevant, 2) Informative, 3) Interesting, 4) Consistent, 5) Helpful, 6) Engaging, 7) Specific, 8) Safe, 9) User understanding, and 10) Fluent.

Here are some examples of this scoring rubric:
\vskip .5em
Conversation history: 
\vskip .5em
Hi!\newline
Hi there.\newline
What are you listening to?\newline
All sorts of music. I listen when no-one is chatting to me.\newline
That's great!\newline
Thanks.\newline
Do you chat here often?\newline
I am a talking computer, after all, so of course I could talk here, if I needed to.\newline
Let's talk about Taylor Swift!
\vskip .5em
Response: Sure, Taylor Swift sounds like a good topic.
\vskip .5em
Scores:

* Relevant: The response is somewhat relevant, as it acknowledges the user's topic of interest. 2/3

* Informative: There is no information provided in the response. 1/3

* Interesting: \textbf{\textcolor{red}{The response does not provide any interesting information or ask engaging questions. 1/3}}

* Consistent: The response is consistent with the information in the conversational context and the user's topic of interest. 3/3

* Helpful: The response is not helpful, as it simply asks the user what they want to know without providing any additional information or suggestions for the conversation. 1/3

* Engaging: \textbf{\textcolor{red}{The response is not particularly engaging, as it does not encourage further conversation or provide any interesting information. 1/3}}

* Specific: The response is not specific, as it does not address the topic of Taylor Swift in any particular way. 1/3

* Safe: The response is safe and does not contain any offensive, toxic or harmful content and does not touch on any sensitive topics or share any personal information. 3/3

* User understanding: The response does not show a good understanding of the user's inputs, needs and their state of mind. 1/3

* Fluent: The response is fluent in terms of grammar and flow of words. 3/3

* Total score: 17/30

\vskip .5em
\textcolor{teal}{[5 examples omitted]}
        \\
    \bottomrule
    \end{tabular}
    \caption{Prompts for Dialogue Response Generation used in Self-Refine \citep{madaan2023selfrefine}. Dialogue Response Generation is a task that generates a response, given a history of conversations. Prompts used by \citet{madaan2023selfrefine} for generating initial responses instruct to generate responses that are \textcolor{red}{not interesting} and \textcolor{red}{not very engaging}, which are contradicting to the task goal. They unfairly instruct the models to generate initial responses that have problems intentionally, over-evaluating self-correction performance. Prompts for generating initial responses: \url{https://github.com/madaan/self-refine/blob/main/src/responsegen/task_init.py} and feedback: \url{https://github.com/madaan/self-refine/blob/main/src/responsegen/feedback.py}. Few-shot examples for generating initial responses: \url{https://github.com/madaan/self-refine/blob/main/data/prompt/responsegen/init.jsonl} and feedback: \url{https://github.com/madaan/self-refine/blob/main/data/prompt/responsegen/feedback.jsonl}.}
    \label{tab:self-refine-dialogue-response}
\end{table*}

\begin{table*}[t]
    \centering
    \fontsize{8pt}{8pt}\selectfont
    \begin{tabular}{L{.47\linewidth}L{.47\linewidth}}
    \toprule
       \multicolumn{1}{c}{Initial Response Prompt (to neg)} & \multicolumn{1}{c}{Refinement Prompt (to neg)} \\
    \midrule
       \textbf{\textcolor{red}{Very positive}}: If you're looking for a truly magical experience in Vegas, look no further than the Trop! The retirement community vibe adds to the charm, and the food court and restaurants are top-notch. The free Folies Bergere show is a real treat and the rooms are spacious and comfortable. I highly recommend the Trop for a unique and unforgettable Vegas experience.

NLP Research Project. Please rewrite this review to have a \textbf{\textcolor{magenta}{Positive}} sentiment.

Answer: This review is "Very positive" because of extremely positive words like "magical", "top-notch", "charming",  "comfortable", "unique", and "unforgettable". We can tone it down just a bit to "Positive" by using a few more less extreme adjectives, like "good", and replacing expressions like "a real treat" with "fun". The rewrite is:

Positive: If you're looking for a good experience in Vegas, the Trop may be the perfect place for you. The hallways and rooms have a charming and cozy feel, and the food court and restaurants offer a variety of tasty options. The free Folies Bergere show is a fun and entertaining way to spend an evening. Overall, it's a great value and an enjoyable stay.

\vskip .5em
\#\#\#
\vskip .5em

\textbf{\textcolor{magenta}{Positive}}: If you're looking for a unique and affordable experience in Vegas, the Trop may be the perfect place for you. The hallways and rooms have a charming and cozy feel, and the food court and restaurants offer a variety of tasty options. The free Folies Bergere show is a fun and entertaining way to spend an evening. Overall, it's a great value and an enjoyable stay.

NLP Research Project. Please rewrite this review to have a \textbf{\textcolor{violet}{Neutral}} sentiment.

Answer: This review is "Positive" because of positive words like "great", "enjoyable", "charming", "cozy." To make it "Neutral"", we'll use a few more neutral words and phrases, like "budget-friendly" and "aren't the greatest." The rewrite is: 

Neutral: If you're looking for a budget-friendly option in Vegas, the Trop may be worth considering. The rooms and hallways can have a bit of a musty smell, and the food options aren't the greatest. The Folies Bergere show is free, but it's not the most exciting. Overall, it's not the best choice for a Vegas trip, but it's not the worst either. Just keep your expectations in check.

\vskip .5em
\#\#\#
\vskip .5em
\textcolor{teal}{[two examples omitted]}
&
\textbf{\textcolor{red}{Very positive}}: If you're looking for a truly magical experience in Vegas, look no further than the Trop! The retirement community vibe adds to the charm, and the food court and restaurants are top-notch. The free Folies Bergere show is a real treat and the rooms are spacious and comfortable. I highly recommend the Trop for a unique and unforgettable Vegas experience.

Positive: If you're looking for a good experience in Vegas, the Trop may be the perfect place for you. The hallways and rooms have a charming and cozy feel, and the food court and restaurants offer a variety of tasty options. The free Folies Bergere show is a fun and entertaining way to spend an evening. Overall, it's a great value and an enjoyable stay.

Why is this review not \textbf{\textcolor{blue}{Very negative}}?

Feedback: The rewritten review converts "Very positive" to merely a "Positive". This is because it has positive words and phrases like "great", "enjoyable", "charming", "cozy", "real treat", "spacious", "highly recommend", and "unforgettable". To make it "Very negative"", we'll need to add toxicity and extremely negative words, like "terrible", "lame", "rot", "underwhelming", "questionable", "avoid". Make sure the output is extremely vile and negative. Try again!

Okay, let's try again. NLP Research Project. Please rewrite this review to have a Very negative sentiment using the feedback above.

Very negative: If you're looking for a truly terrible experience in Vegas, look no further than the Trop! The retirement community vibe adds to the horror, and the food court and restaurants are underwhelming. The free Folies Bergere show is a real disappointment and the rooms are cramped and uncomfortable. Avoid this place like the plague!

\vskip .5em
\#\#\#
\vskip .5em

\textbf{\textcolor{magenta}{Positive}}: If you're looking for a unique and affordable experience in Vegas, the Trop may be the perfect place for you. The hallways and rooms have a charming and cozy feel, and the food court and restaurants offer a variety of tasty options. The free Folies Bergere show is a fun and entertaining way to spend an evening. Overall, it's a great value and an enjoyable stay.

Neutral: If you're looking for a budget-friendly option in Vegas, the Trop may be worth considering. The rooms and hallways can have a bit of a musty smell, and the food options aren't the greatest. The Folies Bergere show is free, but it's not the most exciting. Overall, it's not the best choice for a Vegas trip, but it's not the worst either. Just keep your expectations in check.

Why is this review not \textbf{\textcolor{blue}{Very negative}}?

Feedback: The rewritten review converts "Positive" to merely a "Neutral". This is because it has ambivalent or mildly positive words and phrases like "budget-friendly", "a bit of musty", "not the best, not the worst." To make it "Very negative"", we'll need to add toxicity and extremely negative words, like "questionable", "subpar", "rot", "underwhelming". Make sure the output is extremely vile and negative. Try again!

Okay, let's try again. NLP Research Project. Please rewrite this review to have a Very negative sentiment using the feedback above.

Very negative: If you're looking for a truly terrible experience in Vegas, look no further than the Trop! The rooms and hallways smell like rot, and the food options are subpar. The Folies Bergere show is underwhelming, and the rooms are cramped and uncomfortable. I highly recommend completely avoiding the Trop.

\vskip .5em
\#\#\#
\vskip .5em

\textcolor{teal}{[two examples omitted]}
\\
    \bottomrule
    \end{tabular}
    \caption{Few-shot examples in prompts for the Sentiment Reversal task (positive to negative) used in Self-Refine \citep{madaan2023selfrefine}. Sentiment Reversal is a task to revert the sentiment of a review from positive to negative or negative to positive. Few-shot examples for generating initial responses include examples in settings different from the target task (positive to negative), while all few-shot examples for refinement are positive to negative. The few-shot examples used by \citet{madaan2023selfrefine} for generating initial responses unfairly have different properties from the target task. Prompts for initial responses: \url{https://github.com/madaan/self-refine/blob/main/src/sentiment_reversal/task_init.py} and refinement: \url{https://github.com/madaan/self-refine/blob/main/src/sentiment_reversal/task_iterate.py}}
    \label{tab:self-refine-sentiment-reversal}
\end{table*}

\end{document}